\documentclass[conference]{IEEEtran}
\IEEEoverridecommandlockouts
\usepackage{cite}
\usepackage{hyperref}
\usepackage{amsmath,amssymb,amsfonts}
\usepackage{algorithmic}
\usepackage{graphicx}
\usepackage{subcaption}
\usepackage{textcomp}
\usepackage{xcolor}
\usepackage{enumitem}
\usepackage{booktabs}
\def\BibTeX{{\rm B\kern-.05em{\sc i\kern-.025em b}\kern-.08em
    T\kern-.1667em\lower.7ex\hbox{E}\kern-.125emX}}

\renewcommand{\baselinestretch}{0.9}

\begin{document}

\title{A Transfer Learning Framework for Anomaly Detection in Multivariate IoT Traffic Data
\thanks{This material is based upon work supported by the National Science Foundation under Grant Numbers   CNS- 2318726, and CNS-2232048.} 
}

\author{
\IEEEauthorblockN{Mahshid Rezakhani, Tolunay Seyfi, Fatemeh Afghah}

\IEEEauthorblockA{Holcombe Department of Electrical and Computer Engineering, Clemson University, Clemson, SC, USA \\
        Emails: \{mrezakh, tseyfi, fafghah\}@clemson.edu}
}

\maketitle

\begin{abstract}
In recent years, rapid technological advancements and expanded Internet access have led to a significant rise in anomalies within network traffic and time-series data. Prompt detection of these irregularities is crucial for ensuring service quality, preventing financial losses, and maintaining robust security standards. While machine learning algorithms have shown promise in achieving high accuracy for anomaly detection, their performance is often constrained by the specific conditions of their training data.
A persistent challenge in this domain is the scarcity of labeled data for anomaly detection in time-series datasets. This limitation hampers the training efficacy of both traditional machine learning and advanced deep learning models. To address this, unsupervised transfer learning emerges as a viable solution, leveraging unlabeled data from a source domain to identify anomalies in an unlabeled target domain. However, many existing approaches still depend on a small amount of labeled data from the target domain.
To overcome these constraints, we propose a transfer learning-based model for anomaly detection in multivariate time-series datasets. Unlike conventional methods, our approach does not require labeled data in either the source or target domains. Empirical evaluations on novel intrusion detection datasets demonstrate that our model outperforms existing techniques in accurately identifying anomalies within an entirely unlabeled target domain. 
\end{abstract}


\section{Introduction}

Over the years, industries such as commerce, finance, and healthcare have increasingly relied on systematically collected network traffic data to enhance service quality and detect anomalies. Network traffic data, with its temporal structure, provides a rich source of insights into user behavior, system performance, and potential security threats. By analyzing this data, organizations can identify patterns indicative of normal activity and detect deviations that may signal malicious activity or other anomalies. This focus on network traffic analysis is particularly important in identifying attacks, such as Distributed Denial of Service (DDoS), unauthorized access attempts, or data exfiltration. The temporal sequencing inherent in network traffic data allows for retrospective analysis of past events and serves as a critical input for predictive models that anticipate and mitigate potential future threats. 

Anomaly detection, the task of identifying unusual patterns in data, is increasingly important in environments that require constant monitoring, such as the Internet of Things (IoT). As IoT networks proliferate, they generate vast amounts of sequential data, continuously capturing real-time operational events across smart devices, sensors, and interconnected systems. This data, usually in the form of time-series streams, allows for the tracking of changing conditions in fields like healthcare, industrial automation, and urban infrastructure. However, the complexity of IoT data presents a unique challenge for anomaly detection, as traditional methods often fall short in adapting to the dynamic, continuous, and sequential nature of these streams \cite{IoTAnomaly}.

One key difficulty lies in the characteristic of concept drift, which occurs when the statistical properties of IoT data change over time. Unlike traditional datasets, where data patterns remain stable, IoT data is susceptible to shifts driven by environmental changes, varying device conditions, or even user behaviors. This drift leads to a gradual or sudden change in the data's underlying structure, which, if undetected, can significantly impact the accuracy of anomaly detection models. For instance, sensor data collected from manufacturing equipment may slowly change as the machinery ages, requiring models that can adapt to these evolving patterns. Unfortunately, conventional machine learning approaches, such as decision trees or support vector machines (SVMs), rely heavily on fixed training datasets and static models, making them unsuitable for continuously changing IoT environments \cite{ConceptDrift}.

Traditional machine learning models for anomaly detection typically involve separate training and testing phases. This setup is effective in static conditions, where data patterns remain consistent, but becomes problematic for IoT data. Static models trained on a single dataset are prone to high false-positive or false-negative rates when exposed to new or fluctuating data. For example, a model trained to detect anomalies in sensor data from one device type may struggle to generalize to another due to differing operational characteristics. Such limitations highlight the need for models capable of continuous learning and rapid adaptation to maintain high accuracy and avoid retraining costs—a challenge that is impractical given the high computational demands and limited resources of most IoT devices \cite{ADTechniques}.

One of the most successful architectures for time-series anomaly detection in IoT networks is the autoencoder, particularly those based on Long Short-Term Memory (LSTM) networks. LSTM-based Autoencoders are highly effective for sequential data as they are designed to capture dependencies over time, which is especially useful in identifying deviations from normal patterns in time-series data. These models work by encoding the input data into a lower-dimensional representation and then reconstructing it, with anomalies detected based on the reconstruction error—unusual behaviors result in high errors, indicating deviations from typical patterns. For instance, LSTM-based Autoencoders have been widely used in industrial IoT to identify equipment failures, where the normal operating data pattern changes due to wear or malfunction \cite{UnsupervisedLSTMAD}.

Building on this, Variational Autoencoders (VAEs) extend the capabilities of standard Autoencoders by introducing probabilistic modeling into the anomaly detection process. Unlike conventional Autoencoders, which produce a single deterministic output, VAEs generate a probability distribution over potential outputs, allowing them to better capture the variability of normal data \cite{VAEKingma}. This probabilistic approach enhances the ability to detect subtle anomalies that might not produce significant reconstruction errors but still fall outside the typical data distribution. This is particularly beneficial in unsupervised anomaly detection settings, where labeled data is scarce or unavailable. VAEs are thus well-suited for IoT networks, as they can adapt to new patterns of normal behavior autonomously, an essential feature for rapidly evolving environments \cite{ICASSPVAE}.

In this work, we make several key contributions:
\begin{itemize}[left=0pt, noitemsep]
    \item Construct meaningful sequences based on flows directed to specific receiver IPs, preserving the contextual integrity of network traffic and enhancing anomaly detection accuracy.
    \item Introduce the \textbf{Contrastive Target-Adaptive LSTM-VAE (CTAL-VAE)}, a streamlined architecture that leverages contrastive learning with triplets to effectively capture domain-specific variations while maintaining simplicity and efficiency.
    \item Demonstrate the model’s effectiveness on the target dataset using a few-shot learning approach, where only five triplet ensembles are used for training prior to predicting anomalies on the remaining target data. This highlights a significant achievement, as the model is able to generalize and adapt effectively even with minimal unlabeled data, overcoming the dual challenge of scarce annotations and few-shot scenarios—a critical need in real-world IoT applications.
\end{itemize}

\section{Related Work}
\subsection{IoT Anomaly Detection on Multivariate Time Series Data}

Anomaly detection in IoT systems, especially for multivariate time series data, is critical for ensuring operational resilience, reliability, and security across applications like manufacturing and smart cities. IoT anomaly detection presents unique challenges due to the complexity of multivariate data streams, real-time demands, and limited computational resources on edge devices. The temporal dependencies, diverse sensor types, and high-dimensional feature spaces of IoT data require advanced models that handle spatiotemporal complexities and dynamic environments.

Hybrid approaches are increasingly effective in addressing these challenges. The Real-Time Deep Anomaly Detection (DAD) framework combines Convolutional Neural Networks (CNN) and Long LSTM Autoencoders to capture spatial and temporal features, enabling real-time anomaly detection on resource-constrained edge devices \cite{MultivariateDA}. Similarly, the Adaptive Transformer-CNN (ATC-AFAT) architecture integrates Transformer and CNN modules with adversarial training to improve detection accuracy. Its adaptive attention mechanism enhances temporal sequence reconstruction, while a CNN discriminator strengthens anomaly scoring, making it suitable for industrial IoT applications with minimal latency \cite{TransformerDA}.


The Anomaly Transformer offers an unsupervised approach, leveraging an Anomaly-Attention mechanism to identify association discrepancies between normal and abnormal patterns. It learns prior-associations for adjacent time points and series-associations across the dataset, using a minimax optimization strategy for robust anomaly detection. Its unsupervised nature and ability to handle complex temporal dependencies make it ideal for IoT environments without labeled data \cite{xu2022anomaly}.

\subsection{Transfer Learning for Anomaly Detection}

Flow-based attack detection in IoT systems has gained significant traction due to the ever-increasing deployment of interconnected devices in smart city infrastructures and critical cyber-physical systems (CPS). In \cite{NetworkFlowSriram}, a comprehensive framework is outlined, which employs deep learning techniques to address the limitations of traditional threshold and signature-based methods. Unlike conventional approaches that rely heavily on domain-specific feature engineering and are prone to failure against novel attack variants, deep learning models autonomously extract optimal features from raw network flow data. This capability makes them highly adaptable for identifying botnet traffic in IoT networks.
Complementing this, \cite{ResADMWang} introduces ResADM, a framework that demonstrates the utility of transfer learning for enhancing attack detection in CPS. 
 ResADM achieves remarkable generalization capabilities, maintaining high detection accuracy across different datasets, such as UNSW-NB15 
 and CICIDS2017.
These approaches highlight the importance of integrating flow-based analysis and transfer learning to build robust, adaptable frameworks capable of detecting sophisticated threats in real time. In this context, our focus on data at the receiver side emerges as a logical extension of these strategies, where the aggregation of traffic patterns offers a unique vantage point for identifying anomalous behaviors. Receiver-side analysis not only capitalizes on the comprehensive visibility of network flows but also aligns with our overarching goal of building resilient and scalable detection mechanisms. To accommodate this logic, the datasets are structured into flow-based sequences organized by destination IP, ensuring that traffic data is coherently grouped and analyzed from the receiver's perspective. This method enhances the ability to leverage aggregated traffic patterns, providing a holistic view of network flows to effectively identify and address increasingly complex attack dynamics.

\section{Transfer Learning Using Feature Representation for Multivariate Anomaly Detection}

\begin{figure}[h!]
    \centering
           \vspace{-10pt}
    \includegraphics[width=0.45\textwidth]{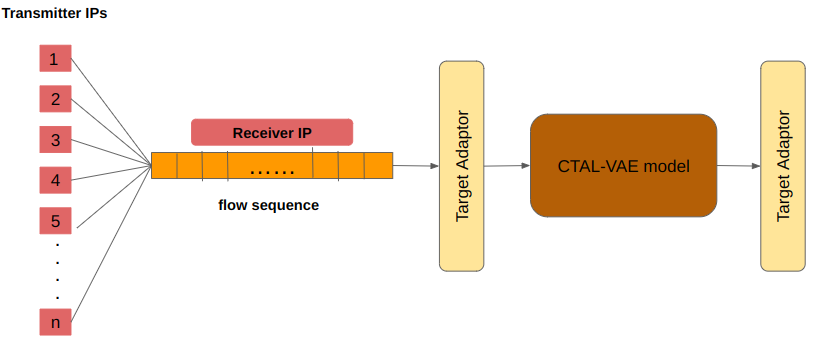}
    \caption{ CTAL-VAE architecture.}
    \label{fig:your_image_labe}
    \vspace{-15pt}
\end{figure}

In this work, we propose the Contrastive Target-Adaptive LSTM-VAE (CTAL-VAE), a novel framework tailored for multivariate anomaly detection in IoT and network environments, as depicted in Figure \ref{fig:your_image_labe}. The CTAL-VAE builds on an LSTM-based Variational Autoencoder architecture, augmented with input and output adaptor layers to bridge the gap between source and target domains. By integrating contrastive learning with supervised triplet ensembles, it enhances the separation of normal and anomalous patterns in latent space, enabling effective adaptation with minimal labeled data. To provide a comprehensive understanding of our approach, we first formalize the problem setting and highlight the unique challenges it poses. Subsequently, we detail the proposed methodology and its components, including the sequence-based VAE architecture, domain-specific adaptors, and the integration of contrastive loss.

\subsection{Problem Definition}


In the context of multivariate time series analysis, data is represented as a sequence of vectors observed at distinct time intervals. Mathematically, this can be expressed as 
\[
\mathbf{X} = (x_1, x_2, x_3, \dots, x_N) \in \mathbb{R}^{m \times n},
\]
where \(m\) represents the length of the series, and \(n\) represents the number of variables. For example, in network traffic analysis, each metric is represented as a single-dimensional time series, tracking its behavior over time. When multiple metrics are analyzed together, they form a multivariate time series, providing a holistic view of the network's performance and behavior across various dimensions.

In our work, we tackle the challenge of detecting anomalies and potential attacks in scenarios where labeled data is scarce. The objective is to build adaptable models capable of leveraging data collected from other environments with differing devices, metrics, and vulnerabilities. These models are designed to operate effectively in new environments with minimal data requirements, enabling robust anomaly detection.

The proposed framework considers two datasets: a source domain \(S\), comprising numerous unlabeled normal flows, and a target domain \(T\), offering only a limited number of normal flows. The approach involves training a model on the source domain to learn an accurate representation of the data and reconstruct it effectively. Once trained, the model is applied to the target domain using a few-shot learning strategy, where only a small number of normal flows from the target are provided. This enables the model to adapt efficiently to the target domain’s characteristics, facilitating reliable anomaly detection with minimal retraining.

\subsection{Methodological Outline}
In IoT networks, Key Performance Indicator (KPI) flows represent multivariate time-series data that often vary in duration, resulting in unevenly spaced data across environments. To address this, data is segmented into equal-sized chunks to create meaningful sequences for model input as depicted in Figure \ref{fig:your_image_labe}. This segmentation establishes a robust baseline for benign behavior, improving the model's anomaly detection capability.

Two key factors shape this approach: (1) IoT networks operate under resource constraints, making real-time processing impractical, and (2) attacks often span extended durations, making it unlikely for a single flow to capture an entire attack. Aggregating flows over time intervals enables the model to analyze both short- and long-term patterns, enhancing detection accuracy.
Instead of relying on fixed time intervals, which could group unrelated flows, we propose constructing sequences based on flows directed to specific receivers. By focusing on traffic patterns destined for a single receiver, the model preserves contextual integrity, improving interpretability and accuracy.

We propose an LSTM-based VAE with two adaptor layers: one at the input to standardize data and another at the output to align predictions with the target domain. These layers address variations between source and target environments, such as sensor types and data formats, ensuring consistent feature representations across domains.
This alignment reduces discrepancies caused by environment-specific differences, allowing the model to learn reliable latent space representations for both domains. By tailoring input and output features, the approach enhances the model's ability to manage the diversity inherent in IoT settings, supporting effective anomaly detection across varied environments.

The VAE-based architecture is central to this approach, providing a robust framework to capture complex sequential dependencies within the IoT data. By using the LSTM's ability to model both short- and long-term patterns, the VAE effectively reconstructs input sequences, allowing it to capture subtle deviations indicative of potential anomalies.
The dual-adaptor design further strengthens the model's sequence reconstruction capabilities, with any deviations between reconstructed and original sequences signaling possible anomalies.

\subsection{Sequence-Based VAE}

The VAE is a deep probabilistic model that learns to encode observed data \( x \) into a latent representation \( z \) while preserving essential features of the data's distribution. This is achieved by modeling the joint relationship between the latent variable \( z \), representing hidden features, and the visible variable \( x \), representing observed data.

In a VAE, the latent variable \( z \) is assumed to follow a prior distribution, typically a standard multivariate Gaussian:
\begin{equation}
    p(z) = \mathcal{N}(z; 0, I).
\end{equation}
The data \( x \) is then generated conditionally on \( z \) through a neural network decoder \( p_\theta(x|z) \). However, the posterior distribution \( p(z|x) \), required to infer the latent variable, is generally intractable. To address this, the VAE introduces an inference network (encoder), which approximates the posterior using a variational distribution:
\begin{equation}
    q_\phi(z|x) = \mathcal{N}(z; \mu(x), \sigma^2(x)),
\end{equation}
where the parameters \( \mu(x) \) and \( \sigma(x) \) are learned through the encoder network and define the latent variable's distribution given the input. The approximate posterior can be expanded as:
\begin{equation}
    q_\phi(z|x) = \frac{1}{\sqrt{2 \pi \sigma^2(x)}} \exp \left( -\frac{(z - \mu(x))^2}{2 \sigma^2(x)} \right).
\end{equation}

To enable gradient-based optimization, the VAE employs the reparameterization trick, where samples of \( z \) are drawn as:
\begin{equation}
    z = \mu(x) + \sigma(x) \cdot \epsilon_{\text{rep}}, \quad \epsilon_{\text{rep}} \sim \mathcal{N}(0, I).
\end{equation}
The decoder then reconstructs the data \( x \) from the sampled \( z \), generating \( \hat{x} \), which approximates the original data \( x \).

In our approach, we consider sequences of network traffic as input to the VAE. Let \(\mathbf{x} = \{x_1, x_2, \ldots, x_T\}\) represent a sequence of network traffic over \( T \) time steps. Instead of treating each \( x_t \) as an independent observation, the VAE models the entire sequence to capture temporal dependencies. The encoder network processes the sequence to approximate the posterior distribution \( q_\phi(\mathbf{z}|\mathbf{x}) \), where \(\mathbf{z}\) is a latent representation summarizing the sequence.

The encoder generates parameters \(\mu_\phi(\mathbf{x})\) and \(\sigma_\phi(\mathbf{x})\) for the latent variable \(\mathbf{z}\),  using a recurrent LSTM to aggregate temporal information:
\begin{equation}
    q_\phi(\mathbf{z}|\mathbf{x}) = \mathcal{N}(\mathbf{z}; \mu_\phi(\mathbf{x}), \sigma^2_\phi(\mathbf{x})).
\end{equation}
Here, \(\mu_\phi(\mathbf{x})\) and \(\sigma_\phi(\mathbf{x})\) are computed from the final hidden state of the LSTM, which summarizes the entire sequence:
\begin{align}
    \mathbf{h}_t &= f_\text{LSTM}(\mathbf{h}_{t-1}, x_t), \\
    \mu_\phi(\mathbf{x}) &= W_\mu \mathbf{h}_T + b_\mu, \quad \log \sigma^2_\phi(\mathbf{x}) = W_\sigma \mathbf{h}_T + b_\sigma,
\end{align}
where \(\mathbf{h}_T\) is the hidden state at the final time step \( T \).

The latent variable \(\mathbf{z}\) is sampled using the reparameterization trick:
\begin{equation}
    \mathbf{z} = \mu_\phi(\mathbf{x}) + \sigma_\phi(\mathbf{x}) \cdot \epsilon_{\text{rep}}, \quad \epsilon_{\text{rep}} \sim \mathcal{N}(0, I).
\end{equation}

The decoder reconstructs the sequence \(\mathbf{x}\) from \(\mathbf{z}\), generating each time step iteratively:
\begin{equation}
    \hat{x}_t = f_\text{decoder}(\mathbf{z}, \hat{x}_{t-1}),
\end{equation}
where \(f_\text{decoder}\) is a recurrent neural network that generates \(\hat{x}_t\) based on \(\mathbf{z}\) and the previously reconstructed step \(\hat{x}_{t-1}\).

This formulation allows the VAE to learn temporal dependencies and behavioral patterns in network traffic, such as the order and timing of flows. By encoding sequences into a latent representation and reconstructing them step-by-step, the model builds a comprehensive understanding of normal traffic behavior, which is essential for identifying subtle anomalies across time.

\subsection{Domain-Specific Adaptors for Variability Handling}

To handle variability between source and target domains, we incorporate \textit{domain-specific adaptor layers} as fully connected layers. These adaptors align target inputs for the encoder and adapt decoder outputs to the target domain, creating a unified feature space for consistent cross-domain learning within the shared VAE architecture. This ensures consistent learning across domains while maintaining domain-specific characteristics in the outputs.

\subsubsection{Encoder Adaptor}

For a sample \( x_{is} \) from the source domain, the \textit{encoder adaptor}, \( A_e \), transforms it into a standardized input \( z_{is} \) suitable for the shared encoder:
\begin{equation}
    z_{is} = A_e(x_{is}).
\end{equation}
Similarly, for a target domain sample \( x_{it} \), the encoder adaptor performs the same transformation:
\begin{equation}
    z_{it} = A_e(x_{it}).
\end{equation}

\subsubsection{Shared Encoder and Decoder Adaptor}

The shared encoder processes the adapted inputs \( z_{is} \) and \( z_{it} \), capturing both sequential dependencies and shared latent representations across domains. Each reconstructed output from the decoder then passes through a \textit{decoder adaptor}, \( A_d \), which restores the domain-specific format. For the source domain, this transformation is given by:
\begin{equation}
    \hat{x}_{is} = A_d(z_{is}).
\end{equation}

\begin{table}[h]
\vspace{5pt}
    \centering
    \caption{Summary of Notation}
    \begin{tabular}{ll}
        \toprule
        \textbf{Symbol} & \textbf{Definition} \\
        \midrule
        \( x_{is} \) & Input sample from the source domain. \\
        \( x_{it} \) & Input sample from the target domain. \\
        \( A_e \) & Encoder adaptor layer aligning domain inputs. \\
        \( A_d \) & Decoder adaptor layer restoring domain-specific outputs. \\
        \( z_{is}, z_{it} \) & Adapted inputs from the source and target domains. \\
        \( \hat{x}_{is}, \hat{x}_{it} \) & Reconstructed outputs for source and target samples. \\
        \( \mu(x), \sigma(x) \) & Parameters defining the approximate posterior distribution. \\
        \( \epsilon_{\text{rep}} \) & Noise sampled from \( \mathcal{N}(0, I) \) for reparameterization. \\
        \bottomrule
    \end{tabular}
    \label{tab:notation}
\end{table}

\subsection{Loss functions and supervised triplet generation}
Our approach to anomaly detection using VAE architecture places considerable emphasis on the design of appropriate loss functions. Our objective is to train the model to accurately represent benign patterns without overfitting. Specifically, the model must balance sensitivity, avoiding both an overreaction to minor benign variations and an indifference to changes indicating anomalies.

VAE traditionally relies on a reconstruction loss, \( \mathcal{L}_{REC} \), and KL divergence loss, \( \mathcal{D}_{KL} \), which drives the model’s focus on faithfully reproducing input data \cite{VAEKingma}. In general, this loss is expressed as:
\begin{equation}
    \mathcal{L}_{REC} = \frac{1}{N} \sum_{i=1}^{N} \| x_i - \hat{x}_i \|^2
\end{equation}
where \( x_i \) represents the input sample, \( \hat{x}_i \) is the reconstructed sample, and \( N \) is the number of samples. We select Mean Squared Error (MSE) as our reconstruction loss to assess the accuracy of the model's reconstructions.

The KL divergence regularizer, \( \mathcal{D}_{KL} \), measures the difference between the true latent variable distribution \( p(z|x) \) and the approximate posterior \( q(z|x) \). For the VAE, it is defined as:
\begin{equation}
\mathcal{D}_{KL}(q_\phi(z|x) || p(z)) = \frac{1}{2} \sum_{j=1}^{d} \big[1 + \log(\sigma_j^2) - \mu_j^2 - \sigma_j^2\big],
\end{equation}
where \( \mu_j \) and \( \sigma_j \) are the parameters of the approximate posterior, and \( d \) is the latent dimension. The KL divergence ensures that the learned latent space aligns with the prior distribution.

To further enhance the separation of latent representations between normal and anomalous data across domains, we apply a contrastive learning approach with supervised triplet generation. Specifically, we generate contrastive pairs by selecting positive and negative examples in relation to an anchor sequence \( x_i \) from each domain according to the approach below:

\begin{itemize}[left=0pt, noitemsep]
    \item \textbf{Positive pairs} \( (x_i, p_i) \): These are created by applying a slight perturbation to the anchor sequence \( x_i \), introducing noise \( \epsilon \) to retain proximity in the latent space.
      \begin{equation}
          p_i = x_i + \epsilon, \quad \epsilon \sim \mathcal{N}(0, \sigma^2)
      \end{equation}
      Here, \( \sigma \) controls the noise level, maintaining similarity between \( x_i \) and \( p_i \).\\
      
    \item \textbf{Negative pairs} \( (x_i, n_i) \): These are generated by selecting a random sample \( n_i \) from a different class than \( x_i \), ensuring a greater distance in the latent space.\\
\end{itemize}

By creating these supervised contrastive pairs, we enable the model to learn distinct latent representations for normal and anomalous data. This approach reinforces the separation in latent space between normal and anomalous sequences, enhancing the model’s ability to differentiate between the two classes across domains.

We chose contrastive loss with cosine similarity over conventional losses like Mean Squared Error (MSE) due to its unique ability to align well with self-supervised objectives by mapping representations to a hypersphere, thereby facilitating effective discrimination across diverse domains \cite{ContrastiveLosses}:
\begin{equation}
\begin{split}
    \mathcal{L}_{\text{CON}} = \frac{1}{N} \sum_{i=1}^{N} \Big[ (1 - y) \cdot (1 - C_{\theta}(x_i, p_i))^2 \\
  + y \cdot \text{max}\big(0, m - (1 - C_{\theta}(x_i, n_i))\big)^2\Big]
\end{split}
\end{equation}
where \( y \) is the label (0 for similar pairs and 1 for dissimilar pairs), \( C_{\theta}(x_i, \{p_i, n_i\}) \) denotes cosine similarity, where \(C_{\theta}(a, b) = \frac{a \cdot b}{\| a \| \| b \|}\),
and \( m \) controls the separation margin between classes.

To balance the contrastive loss \( \mathcal{L}_{\text{CON}} \) with the reconstruction loss, we introduce weighting factors \( \lambda_{REC} \) , \( \lambda_{CON} \), and  \( \lambda_{KL} \), adjusting the impact of each term. Our overall loss function is then expressed as:
\begin{equation}
    \mathcal{L} = \lambda_{CON} \cdot \mathcal{L}_{\text{CON}} + \lambda_{REC} \cdot \mathcal{L}_{REC} + \lambda_{KL} \cdot \mathcal{D}_{KL}
\end{equation}

By optimizing this weighted loss, the model effectively learns intra-class diversity and subtle distinctions within benign data, enhancing the robustness and accuracy of anomaly detection.

\section{Evaluation and Results}
\subsection{Datasets}

In this study, we leverage two distinct IoT datasets from diverse application environments: industrial IoT and military automation IoT, to assess our approach’s effectiveness in anomaly detection within communications and networking contexts.

\subsubsection{WUSTL-IIOT-2021 Dataset} The WUSTL-IIOT-2021 dataset \cite{wustl-2021-paper} originates from an Industrial IoT testbed monitoring water storage tanks. Over 53 hours, 1,194,464 network traffic entries were recorded, including key features like packet drops and flow duration. The dataset is 93\% benign, with 7\% attacks: denial-of-service (89.98\%), reconnaissance (9.46\%), SQL injection (0.31\%), and backdoor (0.25\%). Following prior studies \cite{wustl-2021-paper}, we focus on 23 features most relevant to IIoT intrusion detection.

\subsubsection{ACI-IoT-2023 Dataset} The ACI-IoT-2023 dataset \cite{ACI}, collected in a simulated military IoT setup, represents real-world home automation scenarios. Captured over five days, it contains 742,758 entries, with 95\% attacks and 5\% benign data. Attack types include reconnaissance (74\%), brute force (1\%), and denial-of-service (25\%), covering subcategories like scans and sweeps. This dataset provides insights into IoT vulnerabilities in military environments.

Both the WUSTL-IIOT-2021 and ACI-IoT-2023 datasets represent valid datasets for the field of communications and networking as they capture realistic network traffic scenarios from two key IoT domains: industrial and military automation. The WUSTL-IIOT-2021 dataset provides insights into secure and stable networking requirements in industrial environments, where reliable device communication is essential for operational continuity. In contrast, the ACI-IoT-2023 dataset reflects the networking demands of a military IoT setup, where device interoperability and network resilience are critical due to frequent wireless communication and high vulnerability to diverse attacks. Together, these datasets allow for a comprehensive examination of network behaviors and security needs across varied IoT contexts, reinforcing the role of anomaly detection in enhancing communication security across these environments 

The WUSTL-IIOT-2021 and ACI-IoT-2023 dataset are utilized as the source and target domains, respectively. As described in the proposed method, this approach utilizes unlabeled network traffic flows for both training and testing. The process begins by categorizing the traffic flows directed to each receiver in the network to construct input sequences. From these sequences, 80\% of the benign samples—representing normal data that does not exhibit anomalous behavior—are randomly selected from the WUSTL-IIOT-2021 dataset for training the proposed model and the baseline methods. To adapt the model to the target domain, a transfer learning process is applied, incorporating five representative samples from the target domain sequences. This fine-tuning step ensures the model is optimized for evaluation within the target domain.

\subsection{Few-Shot Learning with Contrastive Training and Target-Specific Adaptors}
The CTAL-VAE is first trained on the source domain to learn general representations of normal behavior. During this phase, the entire VAE is optimized using a contrastive learning approach to enhance the separation of normal and anomalous patterns in the latent space. This ensures that the learned representations are robust and generalizable across domains.
For the target domain, the encoder and decoder adaptors are exclusively activated. During the few-shot training phase on the target data, the core VAE remains frozen, and only the adaptors are updated via gradient-based optimization. Similar to the source training, a contrastive learning approach is employed, enabling the model to adapt effectively to the specific characteristics of the target environment while maintaining the integrity of the representations learned from the source domain.
Few-shot training in our framework involves using a limited number of triplet ensembles from the target domain, where each triplet consists of an anchor, a positive, and a negative sample. In this work, we use 5 shots, corresponding to 5 triplet ensembles, to fine-tune the adaptors. By integrating contrastive loss into both the source training and the target domain's few-shot training, the CTAL-VAE achieves efficient domain adaptation. This dual-phase approach ensures robust anomaly detection, even in environments with minimal labeled or unlabeled data, by leveraging both generalizable source knowledge and target-specific refinements.

\subsection{Evaluation of Model Performance}

\subsubsection{Learning Models and Architecture}
The proposed CTAL-VAE model takes sequences of instances as input and outputs a binary classification. The model, implemented using PyTorch’s neural network module, consists of an encoder, decoder, and hidden layers. Detailed implementation parameters are provided in Table~\ref{tab:model_parameters}. To benchmark performance, a standard autoencoder (AE) and a traditional VAE are also implemented as baseline models. All models are trained on the source dataset and evaluated on the target dataset to assess their performance under domain adaptation.

\begin{table}[h!]
\vspace{5pt}
    \centering
    \caption{Model Architecture and Hyperparameters}
   \resizebox{0.50\textwidth}{!}{ 
        \begin{tabular}{|l|l|}
        \hline
        \textbf{Component} & \textbf{Description / Value} \\ \hline
        Input Adaptor Layer & Fully connected, transforms 78 to 43 \\ \hline
        Encoder LSTM & Input size: 43, Hidden size: 64, Sequence length: 30 \\ \hline
        Latent Space Dimensions & 16 \\ \hline
        Decoder Fully Connected Layer & Transforms latent space to hidden space \\ \hline
        Decoder LSTM & Input size: 43, Hidden size: 64, Sequence length: 30 \\ \hline
        Output Adaptor Layer & Fully connected, transforms 43 to 78 \\ \hline
        Optimizer & Adam \\ \hline
        Learning Rate & 0.001 \\ \hline
        Number of Epochs & 100 \\ \hline
    \end{tabular}}
    \label{tab:model_parameters}
\end{table}

\subsubsection{Evaluation Metrics}
The performance of the proposed model is evaluated using three metrics: Accuracy, Matthews Correlation Coefficient (MCC), and Sensitivity. These metrics provide a comprehensive assessment of the model’s classification capabilities. 
MCC is particularly suited for evaluating performance on imbalanced datasets, capturing the correlation between observed and predicted labels.

\subsubsection{Experimental Results and Analysis}
\begin{figure}[h!]
    \centering
        \vspace{-15pt}
    \includegraphics[width=0.25\textwidth]{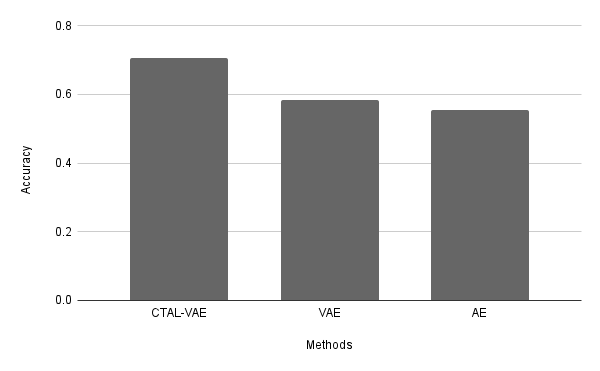}
    \caption{Accuracy comparison across models.}
        \vspace{-10pt}
    \label{fig:accuracy_chart}
    \vspace{-5pt}
\end{figure}

This section summarizes the numerical results of our models for anomaly detection in the target domain. As shown in Figure~\ref{fig:accuracy_chart}, the CTAL-VAE achieves an accuracy of 90\%, outperforming both the VAE and AE models, which attain 82\% and 79\%, respectively. However, accuracy alone is not the most reliable metric for anomaly detection, as models can achieve high accuracy by classifying most samples as normal due to the class imbalance in these datasets.

\begin{figure}[h!]
    \centering
    \begin{subfigure}{0.49\columnwidth}
        \centering
        \includegraphics[width=\textwidth]{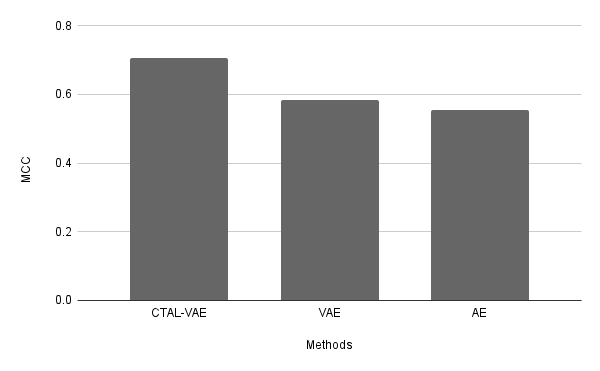}
        \caption{MCC comparison across models.}
        \label{fig:mcc_chart}
    \end{subfigure}
    \hfill
    \begin{subfigure}{0.49\columnwidth}
        \centering
        \includegraphics[width=\textwidth]{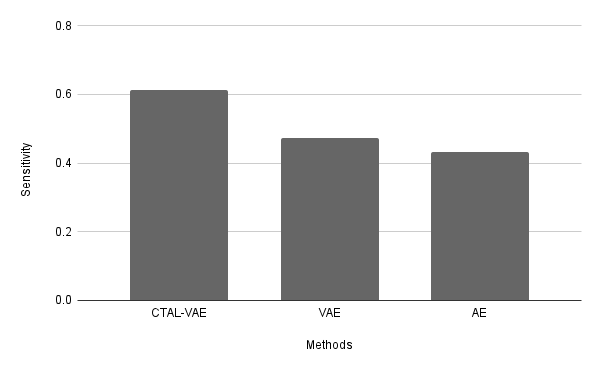}
        \caption{Sensitivity comparison across models.}
        \label{fig:sensitivity_chart}
    \end{subfigure}
    \caption{Comparison of MCC and Sensitivity across models.}
    \label{fig:combined_charts}
     \vspace{-5pt}
\end{figure}

Figure~\ref{fig:mcc_chart} shows the results for the MCC metric, which provides a more balanced assessment of classification performance. The CTAL-VAE achieves the highest MCC score, reflecting its ability to handle imbalanced data effectively. By capturing the correlation between predicted and true labels, MCC demonstrates the robustness of the CTAL-VAE in distinguishing between normal and anomalous instances.

Figure~\ref{fig:sensitivity_chart} presents the sensitivity results, highlighting the model’s capability to detect anomalous conditions. The CTAL-VAE achieves the highest sensitivity among all models, demonstrating its effectiveness in identifying abnormal behaviors, while the AE model shows the lowest sensitivity.

\section{Conclusion}
This work presents the Contrastive Target-Adaptive LSTM-VAE (CTAL-VAE), a novel framework for anomaly detection in multivariate time-series IoT datasets. By combining contrastive learning with a streamlined LSTM-VAE architecture and domain-specific adaptors, the model addresses domain variability and limited labeled data, ensuring robust anomaly detection in dynamic environments. 

Evaluations on WUSTL-IIOT-2021 and ACI-IoT-2023 datasets highlight CTAL-VAE’s superior performance in accuracy, MCC, and sensitivity compared to traditional models. Its ability to generalize across domains with minimal unlabeled data underscores its practical utility.

Future work could explore online learning for continuous adaptation and advanced generative techniques to enhance robustness. Additionally, extending the framework to classify attack types would further strengthen its potential for advanced network security.

\bibliographystyle{IEEEtran}

\begin{thebibliography}{99}

\bibitem{IoTAnomaly}
A. A. Cook, G. Mısırlı, and Z. Fan, ``Anomaly detection for iot time-series data: A survey,'' \textit{IEEE Internet of Things Journal}, vol. 7, no. 7, pp. 6481--6494, 2020. doi: \href{https://doi.org/10.1109/JIOT.2019.2958185}{10.1109/JIOT.2019.2958185}.

\bibitem{ConceptDrift}
L. Yang and A. Shami, ``A lightweight concept drift detection and adaptation framework for iot data streams,'' \textit{IEEE Internet of Things Magazine}, vol. 4, no. 2, pp. 96--101, 2021. doi: \href{https://doi.org/10.1109/IOTM.0001.2100012}{10.1109/IOTM.0001.2100012}.

\bibitem{ADTechniques}
B. Sharma, L. Sharma, and C. Lal, ``Anomaly detection techniques using deep learning in iot: A survey,'' in \textit{2019 International Conference on Computational Intelligence and Knowledge Economy (ICCIKE)}, 2019, pp. 146--149. doi: \href{https://doi.org/10.1109/ICCIKE47802.2019.9004362}{10.1109/ICCIKE47802.2019.9004362}.

\bibitem{UnsupervisedLSTMAD}
O. I. Provotar, Y. M. Linder, and M. M. Veres, ``Unsupervised anomaly detection in time series using lstm-based autoencoders,'' in \textit{2019 IEEE International Conference on Advanced Trends in Information Theory (ATIT)}, 2019, pp. 513--517. doi: \href{https://doi.org/10.1109/ATIT49449.2019.9030505}{10.1109/ATIT49449.2019.9030505}.

\bibitem{VAEKingma}
D. P. Kingma and M. Welling, ``Auto-encoding variational bayes,'' \textit{arXiv preprint arXiv:1312.6114}, 2022. Available at: \url{https://arxiv.org/abs/1312.6114}.

\bibitem{ICASSPVAE}
S. Lin, R. Clark, R. Birke, S. Schönborn, N. Trigoni, and S. Roberts, ``Anomaly detection for time series using vae-lstm hybrid model,'' in \textit{ICASSP 2020 - 2020 IEEE International Conference on Acoustics, Speech and Signal Processing (ICASSP)}, 2020, pp. 4322--4326. doi: \href{https://doi.org/10.1109/ICASSP40776.2020.9053558}{10.1109/ICASSP40776.2020.9053558}.

\bibitem{MultivariateDA}
H. Nizam, S. Zafar, Z. Lv, F. Wang, and X. Hu, ``Real-time deep anomaly detection framework for multivariate time-series data in industrial iot,'' \textit{IEEE Sensors Journal}, vol. 22, no. 23, pp. 22836--22849, 2022. doi: \href{https://doi.org/10.1109/JSEN.2022.3211874}{10.1109/JSEN.2022.3211874}.

\bibitem{TransformerDA}
J. He, Z. Dong, and Y. Huang, ``Multivariate time series anomaly detection with adaptive transformer-cnn architecture fusing adversarial training,'' in \textit{2024 IEEE 13th Data Driven Control and Learning Systems Conference (DDCLS)}, 2024, pp. 1387--1392. doi: \href{https://doi.org/10.1109/DDCLS61622.2024.10606841}{10.1109/DDCLS61622.2024.10606841}.

\bibitem{xu2022anomaly}
J. Xu, H. Wu, J. Wang, and M. Long, ``Anomaly transformer: Time series anomaly detection with association discrepancy,'' \textit{arXiv preprint arXiv:2110.02642}, 2022. Available at: \url{https://arxiv.org/abs/2110.02642}.

\bibitem{NetworkFlowSriram}
S. Sriram, R. Vinayakumar, M. Alazab, and S. KP, ``Network flow based iot botnet attack detection using deep learning,'' in \textit{IEEE INFOCOM 2020 - IEEE Conference on Computer Communications Workshops (INFOCOM WKSHPS)}, 2020, pp. 189--194. doi: \href{https://doi.org/10.1109/INFOCOMWKSHPS50562.2020.9162668}{10.1109/INFOCOMWKSHPS50562.2020.9162668}.

\bibitem{ResADMWang}
H. Wang, H. Zhang, L. Zhu, Y. Wang, and J. Deng, ``ResADM: A transfer-learning-based attack detection method for cyber–physical systems,'' \textit{Applied Sciences}, vol. 13, no. 24, 2023. doi: \href{https://doi.org/10.3390/app132413019}{10.3390/app132413019}.

\bibitem{ContrastiveLosses}
T. Chen, C. Luo, and L. Li, ``Intriguing properties of contrastive losses,'' in \textit{Advances in Neural Information Processing Systems}, vol. 34, pp. 11834--11845, 2021.

\bibitem{wustl-2021-paper}
M. Zolanvari, M. A. Teixeira, L. Gupta, K. M. Khan, and R. Jain, ``Machine learning-based network vulnerability analysis of industrial internet of things,'' \textit{IEEE Internet of Things Journal}, vol. 6, no. 4, pp. 6822--6834, 2019. doi: \href{https://doi.org/10.1109/JIOT.2019.2912022}{10.1109/JIOT.2019.2912022}.

\bibitem{ACI}
N. Bastian, D. Bierbrauer, M. McKenzie, and E. Nack, ``ACI IoT Network Traffic Dataset 2023,'' \textit{IEEE Dataport}, 2023. doi: \href{https://dx.doi.org/10.21227/qacj-3x32}{10.21227/qacj-3x32}.

\end{thebibliography}

\vspace{12pt}

\end{document}